\documentclass{article}

\usepackage{arxiv}

\usepackage[utf8]{inputenc} 
\usepackage[T1]{fontenc}    
\usepackage{hyperref}       
\usepackage{url}            
\usepackage{booktabs}       
\usepackage{amsfonts}       
\usepackage{nicefrac}       
\usepackage{microtype}      
\usepackage{lipsum}

\usepackage{algorithmic}
\usepackage{graphicx}
\usepackage{textcomp}
\usepackage{xcolor}
\usepackage{multirow}
\usepackage{tabularx}

\newcolumntype{C}{>{\centering\arraybackslash}p{2.25em}}

\title{Investigating the Impact of Pre-processing and Prediction Aggregation on the DeepFake Detection Task}

\author{Polychronis Charitidis \quad Giorgos Kordopatis-Zilos \quad Symeon Papadopoulos \quad Ioannis Kompatsiaris \vspace{0.1cm}\\
Information Technologies Institute, CERTH, Thessaloniki, Greece \\
{\tt\small \{charitidis,georgekordopatis,papadop,ikom\}@iti.gr}\\}

\date{}

\begin{document}
\maketitle
\begin{abstract}

Recent advances in content generation technologies (widely known as DeepFakes) along with the online proliferation of manipulated media content render the detection of such manipulations a task of increasing importance. Even though there are many DeepFake detection methods, only a few focus on the impact of dataset preprocessing and the aggregation of frame-level to video-level prediction on model performance. In this paper, we propose a pre-processing step to improve the training data quality and examine its effect on the performance of DeepFake detection. We also propose and evaluate the effect of video-level prediction aggregation approaches. Experimental results show that the proposed pre-processing approach leads to considerable improvements in the performance of detection models, and the proposed prediction aggregation scheme further boosts the detection efficiency in cases where there are multiple faces in a video.

\end{abstract}

\section{Introduction}
\label{Introduction}
The latest advances in synthetic media manipulation have started posing new risks for society and democracy. Although the ability to generate or manipulate facial cues using artificial intelligence could have beneficial applications \cite{forbes,mit} (e.g. art, video games, face anonymization, cinematography), there are several applications that are potentially harmful for individuals, communities and the society as a whole.

The term ``DeepFake'' initially referred to a deep learning-based method able to tamper media by swapping the face between two people. It appeared in 2017 when a machine learning algorithm was employed to transpose celebrity faces into porn videos. Apart from pornography, some of the most harmful usages of this technology include its use for online disinformation and financial fraud \cite{deeptrace}.

\begin{figure}[t]
    \centerline{\includegraphics[width=7cm]{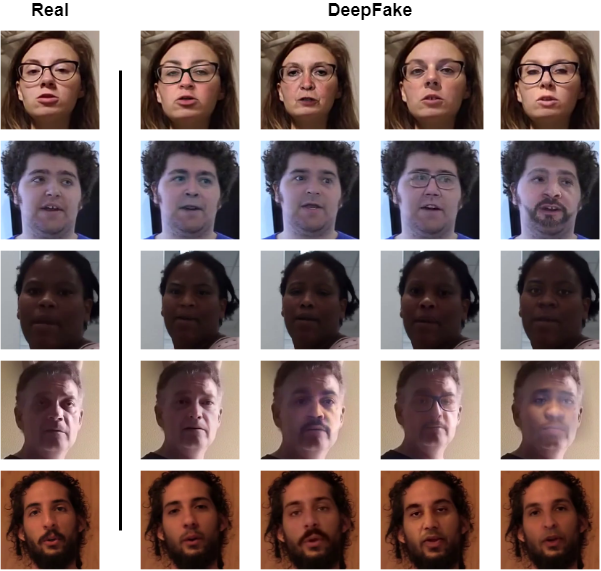}}
    \caption{DFDC dataset examples \cite{dfdc}}
    \label{fig:examples}
\end{figure}

However, ``DeepFakes'' have recently become synonymous with most types of facial and/or audio manipulation. Such manipulations typically include face swap, face generation from scratch, facial attribute manipulation, and facial expression manipulation/reenactment \cite{tolosana2020deepfakes}. Figure \ref{fig:examples} illustrates a few example real faces from the DFDC dataset \cite{dfdc} and their corresponding DeepFake manipulations.

The harmful effects of DeepFakes are widely acknowledged in the research community, and significant effort has been invested lately for the detection of DeepFakes \cite{afchar2018mesonet,rossler2019faceforensics++,dang2020detection,nguyen2019capsule,li2019exposing,cozzolino2018forensictransfer,nguyen2019multi,guera2018deepfake,sabir2019recurrent,amerini2019deepfake}. Several methods leverage recent advances in deep learning,  like the ability to automatically learn rich features with Convolutional Neural Networks (CNNs). The detection problem is typically tackled by training a neural network in a supervised fashion to predict whether an input face is manipulated or not.

Deep learning approaches for DeepFake detection require the availability of large scale datasets. There are numerous such datasets, and the field is progressing rapidly. Besides academic contributions, even large companies like Google, contribute to DeepFake detection research by providing face manipulation datasets \cite{googledataset}. Recently, AWS, Facebook, Microsoft, and the Media Integrity Steering Committee of the Partnership on AI launched the DeepFake Detection Challenge (DFDC) on Kaggle \cite{dfdc}, offering a total prize of \$1,000,000, reflecting the importance that major stakeholders attribute to this problem.

Despite the rapid progress in DeepFake detection and dataset availability \cite{verdoliva2020media}, there has been  very little focus on the pre-processing of training data and how this affects the performance of the resulting detection models. Pre-processing includes all transformations performed on the raw data before they are provided to a model for training or inference. In terms of videos, such transformations often include frame extraction, face detection,  image normalization and resizing, and image augmentation.


In this work, we focus on face detection, an essential step for building accurate DeepFake detectors, as according to \cite{rossler2019faceforensics++}, building detection models around a face results in higher accuracy compared to building models for whole images. A face detector with a large number of false positives will potentially generate a noisy dataset, and this might hurt the overall performance of the detection system. Consequently, a DeepFake detector's performance depends heavily on the performance of the face detection model. After experimenting with various face detectors, we noticed many cases of false positive detections. 
To alleviate this issue, we propose a simple yet efficient pre-processing step that is applied after face extraction and can effectively remove a large number of falsely detected faces. Furthermore, we utilize the information from the proposed approach and develop a prediction aggregation scheme to improve the final video-level prediction and investigate its impact on the detection model compared to other aggregation baselines. Both the proposed data pre-processing scheme and the prediction aggregation improve detection performance significantly on three benchmark datasets.

The work presented in this paper is conducted in the context of the \textit{WeVerify} project \cite{marinova2020weverify}, which aims to build an open-source platform that engages communities and citizen journalists alongside newsroom and freelance journalists for collaborative and decentralised content verification, tracking, and debunking.


\section{Related Work}
\label{Related Work}

\subsection{DeepFake detection}
\label{DeepFake detection}
Since the advent of deep learning, most classification tasks employ deep learning architectures that usually outperform traditional machine learning models. Following this trend, recent DeepFake detection approaches are based on deep learning networks for the detection of manipulated media.

\cite{afchar2018mesonet} present two simple architectures with few layers that exploit mesoscopic features. Meso-4 has four layers of convolutions and pooling followed by a fully-connected layer for classification. Instead, MesoInception-4 is based on a simple variant of the inception module \cite{szegedy2015going}. XceptionNet \cite{chollet2017xception} is proposed as an efficient DeepFake detection architecture \cite{rossler2019faceforensics++}. The same work shows that very deep general-purpose networks outperform shallow CNNs, like MesoInception-4, in the detection task. \cite{dang2020detection} include an attention mechanism to their proposed architecture, which outperformed XceptionNet, when trained on a deepfake dataset they created with various deepfake manipulations. \cite{nguyen2019capsule} presented a capsule-network that requires fewer parameters to train compared with the very deep networks and outperforms shallow nets like MesoInception-4. \cite{li2019exposing} present an approach that exploits artifacts from generated faces with limited resolution, potentially limiting the applicability of the method to specific generator models. 
Other works \cite{yang2019exposing,li2018ictu} detect manipulations utilizing head pose and eye blinking information, respectively; these methods are only evaluated on specific types of manipulations. A more general approach is presented by \cite{cozzolino2018forensictransfer}, where an autoencoder-based architecture is proposed to adapt to new manipulations using just a few examples. This method outperforms XceptionNet in many types of manipulations. The same approach is used by \cite{nguyen2019multi} combining the detection and segmentation tasks to further assist the learning process. This method is promising but as the previous one, requires manipulation masks for training. In  \cite{guera2018deepfake}, a convolutional Long Short Term Memory (LSTM) network is used to exploit temporal dependencies and provide video-level predictions. This work also investigates the impact of the numbers of frames on the final prediction and it reports no further performance gains using more than 40 frames. A recurrent convolutional model was also proposed by \cite{sabir2019recurrent} with competitive results but limited evaluations. \cite{amerini2019deepfake} leverage the optical flow to exploit temporal discrepancies among frames and show that optical flow is predictive of DeepFake manipulations.


\subsection{DeepFake datasets}
\label{DeepFake datasets}
Of the four major categories of DeepFake manipulations, we mainly present the datasets related to face swapping, facial expression manipulation and facial attribute manipulation, as they are the ones we focus on in this work. UADFV \cite{yang2019exposing} is an initial small-scale dataset employing face swapping. The authors in \cite{korshunov2018DeepFakes} present the DeepFakeTIMIT dataset. This consists of 620 fake videos created using a GAN-based face swapping algorithm. FaceForensics++ \cite{rossler2019faceforensics++} is a popular DeepFake dataset that contains 1000 real videos from YouTube. This dataset provides fake videos using face swapping and face reenactment manipulation techniques. The Google/Jigsaw also contributed to the FaceForensics++ dataset with the DeepFake detection dataset (DFD) \cite{googledataset}. The Celeb-DF \cite{li2020celeb} dataset aims to provide face swapping videos of better visual qualities, as previous databases exhibit low visual quality with many visible artifacts. Celeb-DF consists of 408 real videos from YouTube and 795 fake videos. More recently, the DeepFake Detection Challenge (DFDC) \cite{dfdc,dolhansky2019DeepFake} first released a preview dataset consisting of 1131 real videos from 66 paid actors, and 4113 fake videos. The complete DFDC dataset was released on the 11th of December 2019, containing approximately 20,000 real videos and 100,000 fakes. The authors in \cite{jiang2020deeperforensics} present another large-scale DeepFake dataset, including 10,000 fake videos, built using 100 actors.

\subsection{Dataset pre-processing}
\label{Dataset preprocessing}
Although many DeepFake detection works employ data pre-processing, there is very little discussion on the impact of this step on the final detection model. \cite{rossler2019faceforensics++} report that extracting the face region of a video frame, instead of using the whole frame as  input to the deep learning model, yields better detection. Numerous works adopt this approach and use face detection libraries to pre-process the video frames and extract face images. There are many face detection works available \cite{dlib09,zhang2016joint,bazarevsky2019blazeface,deng2019retinaface} and multiple implementations of them, which mainly differ in accuracy, detection speed, and setting availability (e.g., batched detection, detection threshold). In addition to face detection, other works adopt face tracking or face alignment approaches. For example, the authors in \cite{sabir2019recurrent} examine the impact of explicit alignment using facial landmarks and implicit alignment that uses a Spatial Transformer Network (STN) \cite{jaderberg2015spatial} and find that landmark alignment improves the DeepFake detection performance of their presented architecture.


\section{Baseline DeepFake Detection Pipeline}
\label{Baseline DeepFake Detection Pipeline}

\begin{figure*}[t]
    \centerline{\includegraphics[width=15.5cm]{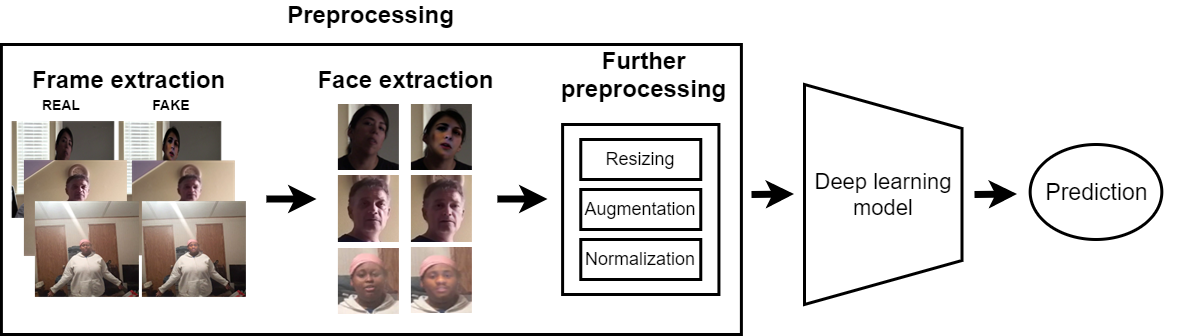}}
    \caption{Baseline DeepFake detection pipeline}
    \label{fig1}
\end{figure*}

Figure \ref{fig1} illustrates a baseline approach for building a video DeepFake detection model. For the training and evaluation of a detection model, one or more of the DeepFake datasets (Section \ref{DeepFake datasets}) can be selected. To transform the raw videos into a format that can be used by deep learning architectures, we apply the three steps depicted in the pre-processing block in Figure \ref{fig1}.

In the first pre-processing step, we extract video frames by applying uniform sampling based on the video duration. Next, we extract the face regions detected in each video frame, 
as recommended by \cite{rossler2019faceforensics++}. 
In this step, the face detectors listed in \ref{Dataset preprocessing} can be used to detect faces and return the face coordinates (in the form of bounding boxes) in the corresponding frames. 
Note that in this step, a common practice is to include some background regions along with the face, as in \cite{rossler2019faceforensics++}. Thus, we multiply the face bounding boxes by a factor of 1.3. The main reason for this is to enable deep learning models to detect resolution inconsistencies or other discrepancies between the face and its surroundings.

The final pre-processing step includes the following transformations: (a) input face image resizing, (b) image augmentation, (c) image normalization, which generally refers to the translation of the RGB pixel values from [0-255] to [0-1]. Such transformations can prevent overfitting in training and generally lead to more robust classifiers.

Using this baseline detection setup, we train state-of-the-art DeepFake detection architectures. The employed architectures
operate at image-level, meaning that we train the models with individual images, and we optimize the model to detect manipulations on them. An additional post-processing step is required to aggregate the per image predictions to a single prediction for the entire video during inference. We experiment with different post-processing aggregation methods for video-level predictions, and we investigate an aggregation approach that utilizes the concept of the connected components according to the proposed pre-processing step (Section \ref{Proposed Preprocessing Approach}).

\begin{figure}[t]
    \centerline{\includegraphics[width=\columnwidth]{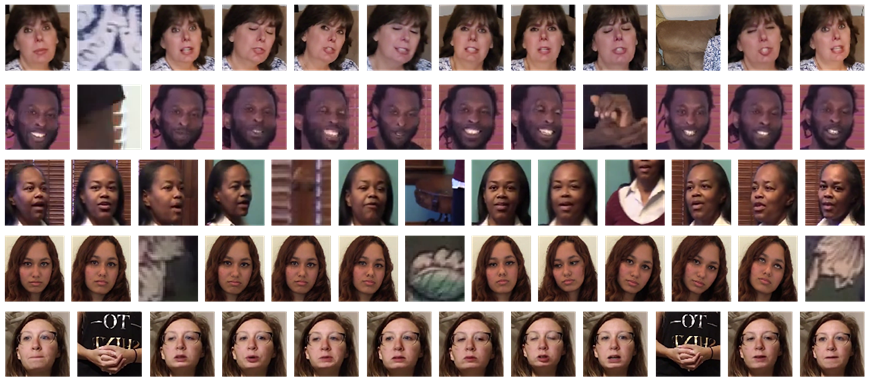}}
    \caption{Extracted face detection regions from random DFDC videos (rows). Among the detected faces, there are several cases of false positive detection.}
    \label{fig2}
\end{figure}

The issue with the pre-processing pipeline of Figure \ref{fig1} is that it depends on the accurate face detection for the generation of face images. Having experimented with multiple face detectors, we noticed that the amount of false positives is higher than expected. Of course, one could fine-tune the detection settings (e.g., increase the detection confidence threshold) to minimize false detections, but this process is time-consuming, depends on the examined dataset and runs the risk of removing correctly detected faces. Using the default settings of a publicly available implementation for face detection \cite{zhang2016joint}, we extracted the face regions from some random DFDC videos. The results are presented in Figure \ref{fig2}. One may notice that among correctly detected faces, there are cases where the face detector failed to detect a human face. False detections usually include random shapes, various human body parts (e.g., hands, neck) and regions with a small proportion of the face displayed. Additionally, we empirically found that false detections are usually not consistent across the duration of videos, meaning that they do not appear in every extracted video frame.

However, 
false detections can lead to noisy training data, which could impact the DeepFake detection performance. 
To address this issue, we propose an additional pre-processing step to clean the dataset by removing detected images that do not contain faces.

\section{Proposed Pre-processing Approach}
\label{Proposed Preprocessing Approach}

Here, we describe the proposed pre-processing step for removing false face detections, 
which we apply after the face detection step (Figure \ref{fig1}).

\begin{figure*}[t]
    \centerline{\includegraphics[width=12cm]{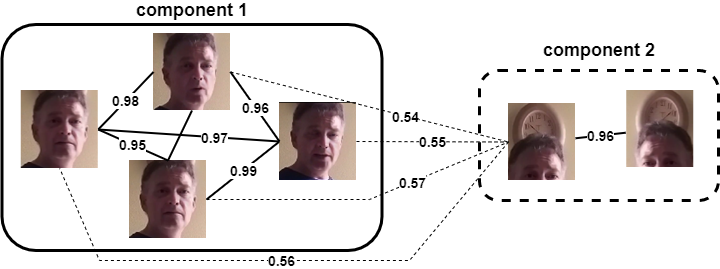}}
    \caption{The proposed pre-processing step. Images that are similar to each other (solid lines) form connected components. Dashed lines connect images with low similarity. For simplicity, we show only a subset of possible connections among faces. Components of small size (cf. section \ref{Method description}) 
    (dashed border) are removed.}
    \label{fig3}
\end{figure*}

\subsection{Method description}
\label{Method description}
The main objective of the proposed step is to generate clusters with correct and incorrect detections, i.e., face and non-face images, so as to remove the latter. We have noticed that false detections occur randomly throughout the video, and they are not repeated in every frame. Furthermore, faces should be present throughout the video, which is valid for several DeepFake datasets. Therefore, clusters formed by incorrect detections should have a much smaller size than clusters of correctly detected faces. 

To this end, we employ a face recognition model, based on the work by \cite{schroff2015facenet}, in order to compute facial embeddings for the extracted images. Embeddings encode the facial information in $D$-dimensional vectors. The employed architecture extracts embeddings with $D=512$. To this end, we calculate the similarity between the detected faces based on the dot product of the corresponding embedding vectors. More formally, let $i,j \in \{1, 2,  \cdots, K\}$ where $K$ is the number of detected faces in $N$ extracted frames, 
then the similarity between the $i$-th and $j$-th detected face is defined as: 
\begin{equation}
\label{eq:1}
s(i,j) = f_w(i)^\top \cdot f_w(j)
\end{equation}
where $f_w(\cdot)$ is the embedding function that maps an arbitrary face image to the facial embedding space $\mathbb{R}^D$, and $s(\cdot,\cdot)$ is the function that assesses the similarity between two face images.

Then, we utilize the similarity information between the detected faces to generate connected components on a \textit{face graph}. The graph nodes correspond to the detected face images. For the face graph formation, two nodes $i$, $j$ are connected with an edge if their similarity is greater than a predefined threshold $\theta$, i.e., $s(i,j) > \theta$. We experimentally validate the impact of the similarity threshold $\theta$ in evaluation datasets in Section \ref{Impact of selected thresholds}.

\begin{figure*}[t]
    \centerline{\includegraphics[width=\textwidth]{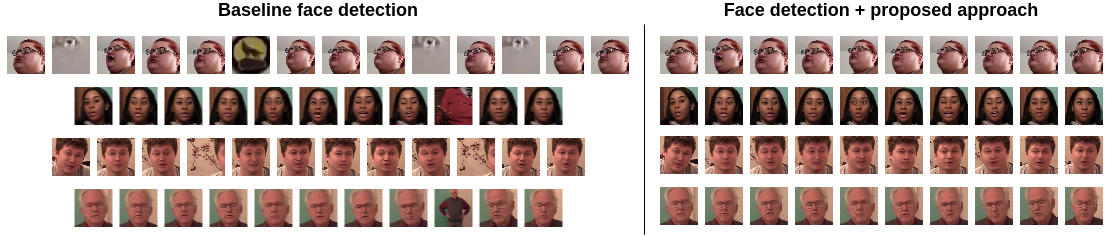}}
    \caption{Qualitative results from DFDC videos. Faces extracted with baseline preprocessing (left). Faces extracted adding the proposed preprocessing step (right). Frames sampled with 1 frame per second.}
    \label{fig:qualitatave}
\end{figure*}

Figure \ref{fig3} illustrates this process. Nodes with similarity greater than $\theta=0.8$ are connected with each other (solid line); otherwise, there is no edge between nodes (dashed line). After this process is completed, if there are no false detections, we expect the number of the generated components to be equal to the number of distinct faces in a video. In cases where there are false detections, these should form an independent component, i.e., Component 2 in Figure \ref{fig3}. As mentioned above, these components will usually contain fewer nodes in comparison to the ones formed by correct detections. This is valid in most cases because the face detectors produce false results in only a subset of the frames. We have empirically found that removing components with size less or equal than $N_F/2$, where $N_F$ is the number of the sampled frames of a video containing at least one detected face, leads to better dataset quality.
%
We qualitatively checked 50 random videos from the DFDC dataset. Figure \ref{fig:qualitatave} depicts some of the qualitative results. Furthermore we investigate the impact of the component size threshold in evaluation datasets in Section \ref{Impact of selected thresholds}.
In the provided example in Figure \ref{fig3}, let $N_F=4$\footnote{Note that in this particular example, some of the detected faces may come from the same frame and this is why $N_F$ is smaller than the number of nodes in the face graph.}, then Component 2 with only 2 nodes is removed, and only images from component 1 are forwarded to the next step of the pre-processing pipeline.

\subsection{Advantages and limitations}
\label{Advantages and limitations}

The main advantages of the proposed approach are that it is a simple way to increase detection performance, and it is very fast. Also, it can be used on top of any existing face detection process, in combination with any available face detection library.


Furthermore, the facial embedding information can be utilized in order to separate the detected faces to clusters of different persons. This functionality is not available by most face detection implementations. 
This is particularly useful for making separate predictions per face (person), especially for cases where there is only one manipulated face among many in a video. This technique is invariant to face movements and it can accurately track multiple moving faces and form separate face clusters.

Our approach has two main limitations. The first is that it assumes that faces are present throughout the duration of a video. So, in cases where a person appears only for a small fraction of the video duration, this face will likely be considered a false detection and consequently will be removed. Even though this is generally not the case in the DeepFake datasets, it can be encountered in several real-world online manipulated videos. The second limitation is that we consider clusters of small sizes to be false detections, and although this is usually the case, there are cases where the face detector can make the same incorrect detections in every extracted frame. In that case, the resulting cluster will be large and hence our approach will incorrectly assume that it corresponds to a correctly detected face. 

\section{Experimental Study}
\label{Experimental study}

\subsection{Pre-processing setup}
\label{Preprocessing setup}

To examine the impact of pre-processing on the detection accuracy, we evaluate two different pre-processing approaches. The first approach, which we denote as baseline, is based on the pipeline described in Section \ref{Baseline DeepFake Detection Pipeline}. The second approach additionally contains the proposed pre-processing step, as described in Section \ref{Proposed Preprocessing Approach}, after the face detection. 

For both pre-processing approaches, we use the method proposed in \cite{zhang2016joint} to detect faces, setting the face detection threshold to 0.9, as a good trade-off between reducing false detections and not increasing false negatives. 
We also expand the size of a detected bounding box by a factor of 1.3, as reported in \cite{rossler2019faceforensics++}. Additionally, for model training, we apply several augmentations on the extracted images: 
horizontal and vertical flipping, random cropping, rotation, compression, Gaussian and motion blurring, and brightness, saturation, and contrast transformation\footnote{For implementation, we made use of the \texttt{imgaug} Python library, \url{https://imgaug.readthedocs.io/en/latest/index.html}}. Also, we normalize the input RGB values to the [0-1] interval. For ImageNet pre-trained models, i.e., XceptionNet and EfficientNet-B4, we employ the normalization scheme they have been trained with, including the channel-wise subtraction of the ImageNet dataset mean, followed by division with the ImageNet dataset standard deviation. In the rare case where no faces are detected in a video during training, we do not further consider the video, while during the inference phase, we manually set the prediction to 0.5.

\subsection{Training setup}
\label{Training setup}

We use the DFDC dataset for model training. The dataset contains approximately 20,000 real videos and 100,000 fakes. Of those, we use 1000 videos for validation, 1202 for testing, and the rest for training. The validation and test sets have equal number of real and fake videos. We also generate two face datasets with these videos using both the baseline and the proposed pre-processing approach. For the proposed training dataset generation, we apply our pre-processing approach to all frames of the video in order to extract faces.
To deal with the class imbalance, we sample 16 of these faces from the real videos and only 4 from the fake ones during batch generation. We randomly sample the same amount of frames every epoch from each video to increase training diversity. For the baseline pre-processing dataset, we sample the same number of detected faces.

We experiment with three different deep learning architectures: MesoInception-4 \cite{afchar2018mesonet}, which was designed specifically for DeepFake detection, XceptionNet \cite{chollet2017xception}, which outperforms other deep networks \cite{rossler2019faceforensics++}, and EfficientNet \cite{tan2019efficientnet} (the EfficientNet-B4 variant), which achieves state-of-the-art performance in most image classification tasks. To adapt the last two architectures to the DeepFake setting, we employ the corresponding backbone networks and add two fully connected layers with 512 and 1 neurons, respectively. We use the sigmoid activation function in the final layer for classification. Except for MesoInception-4, the other models are initialized using the ImageNet pre-trained weights \cite{russakovsky2015imagenet}. For training, we use the Adam optimizer \cite{kingma2014adam} and minimize the Log loss error. Note that the training process operates at the image and not at video level. We train the networks for 10 epochs with a learning rate of $10^{-4}$, and with batch size of 84 for MesoInception-4 and 16 for XceptionNet and EfficientNet. We select the model with the best validation error for each architecture.

\subsection{Evaluation setup}
\label{Evaluation setup}

To examine the impact of the different preprocessing approaches on the trained detection models we evaluate them on the Celeb-DF \cite{li2020celeb}, FaceForensics++ \cite{rossler2019faceforensics++} datasets and DFDC test datasets. Celeb-DF consists of 408 real and 795 fake videos. FaceForensics++ consists of 1000 real videos and 4000 fake videos. To balance datasets, we randomly subsample the majority class in the case of Celeb-DF. For FaceForensics++, we use 1000 real videos and only 1000 fake videos with DeepFakes manipulation, ignoring the other three manipulations (Face2Face, FaceSwap, Neural Textures). Since the training dataset and DFDC test dataset originate from the same distribution of videos, i.e., the same face manipulations were used, we expect detection performance in this dataset to be significantly better in comparison to the other two evaluation datasets.


For evaluation, we extract 4 frames per second and we run experiments using the detection models that were trained with the baseline and proposed pre-processing approaches. 
For the proposed pre-processing approach, we make separate predictions for every detected face image in a video. To aggregate these predictions, we consider four approaches. 
a) averaging the individual predictions (\textit{Avg}),
b) taking the median prediction (\textit{Median}), 
c) taking the maximum prediction (\textit{Max}),
d) averaging predictions per face using the component information from the pre-processing step and taking the maximum prediction among faces (\textit{Face}).
To measure detection performance, we report the aggregated video-level Log loss error and accuracy for each setting.

\begin{table*}[htbp]
\caption{Log loss error, Accuracy (\%) and F1 score of three architectures on three DeepFake datasets. For comparison, note that the Log loss error of the dummy classifier (always predict 0.5) is 0.693.}
\begin{center}
\begin{tabular}{c}
    \scalebox{0.75}{
        \begin{tabular}{|c|c||C|C|C|C||C|C|C|C||C|C|C|C|}
            \hline
            \multirow{2}{*}{\textbf{Model}}&\textbf{Pre-}&\multicolumn{4}{c||}{\textbf{Celeb-DF}} &\multicolumn{4}{c||}{\textbf{FaceForensics++}}&\multicolumn{4}{c|}{\textbf{DFDC test}}\\ 
            \cline{3-14} 
            & \textbf{processing} & \textbf{\textit{Avg}}& \textbf{\textit{Med.}} & \textbf{\textit{Max}}& \textbf{\textit{Face}} & \textbf{\textit{Avg}}& \textbf{\textit{Med.}} & \textbf{\textit{Max}}& \textbf{\textit{Face}} & \textbf{\textit{Avg}}& \textbf{\textit{Med.}} & \textbf{\textit{Max}}& \textbf{\textit{Face}}\\
            \hline
            \hline
            \multirow{2}{*}{MesoInc4} & baseline & 0.677 & 0.689 & 0.782 & 0.679
                                                        & 0.670 & 0.680 & 0.769 &  0.672
                                                        & 0.484 & 0.491 & 0.540 & 0.463 \\
            \cline{2-14} 
            & proposed & 0.644 & 0.657& 0.791 & \textbf{0.642} &
                       \textbf{0.635} & 0.639 & 0.705& 0.636
                       &0.420 & 0.441 & 0.503 & \textbf{0.401} \\
            \hline
            \hline
            \multirow{2}{*}{Xception} & baseline & 0.562 & 0.573 & 0.601 & 0.561 
                                                    & 0.584 & 0.598 & 0.613& 0.585 
                                                   & 0.354 & 0.364 & 0.402& 0.333  \\
            \cline{2-14} 
             & proposed & \textbf{0.520} & 0.530 & 0.549& 0.522& 
                        \textbf{0.543} & 0.546 & 0.580& 0.544
                        & 0.312 & 0.324 & 0.382& \textbf{0.292}  \\
            \hline
            \hline
            \multirow{2}{*}{EffNet-B4}& baseline & 0.510 & 0.519 & 0.569& 0.508 
                                                        & 0.568 & 0.578 & 0.601& 0.566 
                                                        & 0.213 & 0.223 & 0.310&  0.198\\
            \cline{2-14} 
            & proposed & 0.463 & 0.495 & 0.558 & \textbf{0.453}
                        & \textbf{0.497} & 0.515 & 0.621& 0.499 
                        & 0.195 & 0.243 & 0.320& \textbf{0.173} \\
            \hline
        \end{tabular}
        \label{tab1}
    }\\
    \small (a) Log loss error \normalsize
    \\
    \scalebox{0.75}{
        \begin{tabular}{|c|c||C|C|C|C||C|C|C|C||C|C|C|C|}
            \hline
           \multirow{2}{*}{\textbf{Model}}&\textbf{Pre-}&\multicolumn{4}{c||}{\textbf{Celeb-DF}} &\multicolumn{4}{c||}{\textbf{FaceForensics++}}&\multicolumn{4}{c|}{\textbf{DFDC test}}\\ 
            \cline{3-14} 
            & \textbf{processing} & \textbf{\textit{Avg}}& \textbf{\textit{Med.}} & \textbf{\textit{Max}}& \textbf{\textit{Face}} & \textbf{\textit{Avg}}& \textbf{\textit{Med.}} & \textbf{\textit{Max}}& \textbf{\textit{Face}} & \textbf{\textit{Avg}}& \textbf{\textit{Med.}} & \textbf{\textit{Max}}& \textbf{\textit{Face}}\\
            \hline
            \hline
            \multirow{2}{*}{MesoInc4} & baseline & 67.8 & 67.5 & 61.2 & 67.2
                                                        & 66.7 & 66.4 & 60.2& 66.4
                                                        &75.8 & 75.7 & 64.0 & 76.9 \\
            \cline{2-14} 
            & proposed & 69.3 & 69.1 & 61.2 &\textbf{69.6}&
                       \textbf{67.2} & 67.0 & 59.8& 66.8
                       &76.6 & 76.5 & 64.4 & \textbf{78.4} \\
            \hline
            \hline
            \multirow{2}{*}{Xception} & baseline & 76.8 & 76.7 & 63.0& 76.8& 
                                                75.4 & 75.1 & 67.8 & 75.2
                                                & 85.5 & 85.0 & 69.1& 85.8  \\
            \cline{2-14} 
             & proposed & 78.5 & 78.2 & 66.3& \textbf{78.6}& 
                        \textbf{77.2} & 76.0 & 68.1 & 77.0
                        & 86.1 & 87.7 & 70.1& \textbf{87.8}  \\
            \hline
            \hline
            \multirow{2}{*}{EffNet-B4}& baseline & 81.8 & 81.4 & 72.1 & 81.6
                                                        & 78.4 & 78.1 & 69.1 & 78.9
                                                        & 92.1 & 92.1 & 84.3 & 93.8 \\
            \cline{2-14} 
            & proposed  & 83.0 & 82.9 & 71.0 & \textbf{83.1}
                        & \textbf{81.2} & 80.8 & 70.2 & 81.0
                        & 94.2 & 94.5 & 84.9 & \textbf{96.3} \\
            \hline
        \end{tabular}
        \label{tab2}
    }\\
    \small (b) Accuracy \normalsize
    \\
    \scalebox{0.75}{
        \begin{tabular}{|c|c||C|C|C|C||C|C|C|C||C|C|C|C|}
            \hline
           \multirow{2}{*}{\textbf{Model}}&\textbf{Pre-}&\multicolumn{4}{c||}{\textbf{Celeb-DF}} &\multicolumn{4}{c||}{\textbf{FaceForensics++}}&\multicolumn{4}{c|}{\textbf{DFDC test}}\\ 
            \cline{3-14} 
            & \textbf{processing} & \textbf{\textit{Avg}}& \textbf{\textit{Med.}} & \textbf{\textit{Max}}& \textbf{\textit{Face}} & \textbf{\textit{Avg}}& \textbf{\textit{Med.}} & \textbf{\textit{Max}}& \textbf{\textit{Face}} & \textbf{\textit{Avg}}& \textbf{\textit{Med.}} & \textbf{\textit{Max}}& \textbf{\textit{Face}}\\
            \hline
            \hline
            \multirow{2}{*}{MesoInc4} & baseline & 64.2 & 63.0 & 59.5 & 64.6
                                                        & 63.9 & 64.2 & 58.2& 64.2
                                                        & 74.2 & 74.0 & 63.3 & 76.2 \\
            \cline{2-14} 
            & proposed & 67.0 & 66.1 & 62.2 &\textbf{67.6}&
                       65.2 & 64.9 & 58.7& \textbf{66.1}
                       &77.1 & 76.5 & 65.4 & \textbf{78.8} \\
            \hline
            \hline
            \multirow{2}{*}{Xception} & baseline & 75.5 & 74.9 & 60.1& 75.3& 
                                                73.9 & 72.8 & 67.2 & 73.5
                                                & 86.2 & 85.5 & 69.9& 86.3  \\
            \cline{2-14} 
             & proposed & 77.2 & 76.9 & 62.3& \textbf{77.6}& 
                        \textbf{75.8} & 74.2 & 66.9 & 75.2
                        & 87.2 & 86.8 & 73.1& \textbf{88.2}  \\
            \hline
            \hline
            \multirow{2}{*}{EffNet-B4}& baseline & 82.7 & 81.4 & 75.1 & 82.5
                                                        & 77.6 & 76.1 & 73.2 & 77.0
                                                        & 93.0 & 92.5 & 83.5 & 94.3 \\
            \cline{2-14} 
            & proposed  & 83.1 & 82.5 & 74.0 & \textbf{84.1}
                        & \textbf{79.2} & 78.8 & 73.2 & 78.4
                        & 95.1 & 94.5 & 83.9 & \textbf{96.5} \\
            \hline
        \end{tabular}
        \label{tab3}
    }\\
    \small (c) F1 score \normalsize
    \\
\end{tabular}
\end{center}
\end{table*}
\subsection{Experimental results}
\label{Results}

Table \ref{tab1}(a) illustrates the Log loss error of the three benchmarked models on every evaluation dataset. Note that the error for a dummy classifier that always predicts 0.5 for each video is 0.693. For the case of the Celeb-DF dataset, we notice that EfficientNet-B4 achieves the best performance among all models. MesoInception-4 marginally surpasses the performance of the dummy classifier. This indicates that shallow architectures are not suitable for the DeepFake detection task. Models trained with the proposed pre-processing approach outperform the baseline by a significant margin. For example, the performance gain of the models with \textit{Avg} aggregation ranges between 0.033 and 0.055 in terms of Log loss error, corresponding to 5-11\% relative performance increase. In terms of the aggregation methods, the \textit{Face} and \textit{Avg} methods achieve the best results exhibiting similar performance, with the former achieving marginally better results. This is expected because videos in the Celeb-DF dataset contain one face throughout the video, and as a result, both approaches average the predictions, except for some cases where the pre-processing approach detects multiple components. The \textit{Median} approach is slightly worse than the previous two, and \textit{Max} achieves the worst performance.

Similar conclusions apply for the FaceForensics++ dataset. EfficientNet-B4 with the \textit{Avg} aggregation achieves the best performance. Once again, it is clear that pre-processing is beneficial for the performance of the detection models, even more than in the Celeb-DF dataset. In almost all model-aggregation combinations, the Log loss error drops considerably when the network is trained with the proposed approach compared to the baseline. For example, the performance gain of the models with \textit{Avg} ranges between 0.035 and 0.067, corresponding to a 5-12\% relative performance increase. Also, \textit{Face} and \textit{Avg} aggregations demonstrate similar performance, which can be attributed to the fact that the FaceForensics++ dataset contains one face per video as well.

The Log loss error of all models on the DFDC test is significantly smaller compared to other datasets.
This is expected because the test set contains videos with the same manipulations as the training set. 
It is noteworthy that the models trained with the proposed pre-processing approach consistently outperform those trained with the baseline. In this case, models with the proposed pre-processing score 12-13\% better. Additionally, the \textit{Face} aggregation scheme outperforms the \textit{Avg} by a clear margin. This can be attributed to the fact that the DFDC dataset contains several videos with more than one person. Also, only one of the faces in these videos are manipulated, which would render \textit{Avg} an almost random way of aggregating individual predictions. 

Table \ref{tab1}(b) displays the results in terms of accuracy. We derive similar conclusions as in the case of Log loss error. In general, accuracy is improved when training the models with the proposed pre-processing pipeline. The average gain for \textit{Face} and \textit{Avg} aggregation methods is approximately 2\%. The \textit{Face} aggregation scores slightly better than \textit{Avg} for the  Celeb-DF and the opposite for the FaceForensics++ dataset. The \textit{Face} aggregation method outperforms \textit{Avg} by a larger margin for the case of DFDC test dataset. 

For completeness, we also report the macro-average F1 score in Table \ref{tab1}(c). Macro-average F1 improved when training the models with the proposed pipeline. 
Reporting the best aggregation approach per dataset, we observe a 2.3\% performance increase with the proposed pre-processing for the Celeb-DF dataset. This value is 1.6\% and 2.2\% for the FaceForensics++ and DFDC test datasets respectively.


\subsection{Impact of selected thresholds}
\label{Impact of selected thresholds}


In this section, we investigate the impact of the similarity threshold and component size threshold on detection performance. We pre-process evaluation datasets with the proposed pre-processing approach using different similarity thresholds and employ the detection models to evaluate them. We also investigate the impact of the component size threshold on the model's performance testing three different threshold values on the evaluation datasets. 
We experiment only with the EfficientNet model with the \textit{face} aggregation method.

Table \ref{tab5} displays the Log loss error of our model for three similarity thresholds on three test datasets. It is evident that using 0.8 similarity threshold yields the best results for the case of FaceForensics++ and DFDC test datasets. The best performance in Celeb-DF is achieved with 0.9 similarity threshold. Generally, the performance difference between 0.8 and 0.9 similarity thresholds is marginal. On the other hand, the performance drop is more significant when the similarity threshold is lower ($\theta\le$ 0.7). This is expected as lowering the similarity threshold leads to generated components with more dissimilar faces. This means that more false detections are connected with true detections to form a component, limiting the dataset cleaning capabilities.

Table \ref{tab5} also shows the experimental results when applying the proposed pre-processing with different minimum component size threshold. The similarity threshold for these experiments is 0.8. In the cases where all components are removed for a video, then we set the final video prediction to 0.5. When we increase the minimum size threshold to $3N_F/4$, then the evaluation error is higher. This happens because there are cases in the dataset where a component with correctly detected faces is removed, leading to no generated components for a particular video. On the other hand, we observe a measurable increase in the error when the size threshold decreases to $N_F/4$. For the case of the Celeb-DF dataset, the error remains the same as the initial one, meaning that there is no change in the detected components. In the other two evaluation datasets, a small increase in the error is observed, owing to the fact that the proposed process fails to remove some incorrectly detected components in certain videos.
\begin{table*}[t]
\caption{Log loss error of the EfficientNet-B4 architecture with different similarity thresholds and varying component sizes on three DeepFake datasets}
\begin{center}
    \scalebox{0.88}{
    \begin{tabular}{|l||c|c|c||c|c|c|}
        \hline
        \multirow{2}{*}{\textbf{Dataset}} & \multicolumn{3}{c||}{\textbf{Similarity threshold}} & \multicolumn{3}{c|}{\textbf{Size threshold}} \\
        \cline{2-7} 
        & \textit{0.7} & \textit{0.8} & \textit{0.9} & $N_F/4$ & $N_F/2$ & $3N_F/4$ \\ \hline\hline
        \textbf{Celeb-DF}  & 0.480 & 0.453 & \textbf{0.445} & \textbf{0.453} & \textbf{0.453} & 0.475\\ \hline
        \textbf{FaceForensics++}      & 0.525 & \textbf{0.499} & 0.506 & 0.510 & \textbf{0.499} & 0.544\\ \hline
        \textbf{DFDC test} & 0.204  & \textbf{0.173} & 0.194   & 0.182 & \textbf{0.173} & 0.230\\ \hline
    \end{tabular}
    \label{tab5}
    }
\end{center}
\end{table*}

\section{Discussion and Conclusions}
\label{Discussion and Conclusions}

In this work, we studied the impact of dataset pre-processing and prediction aggregation on the DeepFake detection problem. We proposed a pre-processing step that improves the training dataset quality. We found that pre-processing has a considerable impact on the detection task and boosts model performance by improving the quality of the generated training set. Using the information derived from this approach, we experimented with an aggregation approach that outperforms baselines when multiple faces appear in a video.

A noteworthy observation of our work is the lack of detection generalization in unseen manipulations. It is apparent that when the detection models are trained with a dataset of specific facial manipulations, then the performance on other datasets with different manipulations is poor. This means that research efforts have to focus on the generalization of the detection models, especially now that new manipulation methods appear by the day. Additionally, the broad definition of DeepFake manipulations that includes all kinds of facial manipulation and face generation makes the problem even harder.

In the future, we plan to further optimize the performance of our Deepfake Detection pipeline and deploy it as a web service and make it accessible to journalists and citizens through the web media forensics tool \cite{zampoglou2016web} and the InVID-WeVerify verification plugin \cite{teyssou2017invid}. In that way, we will explore both challenging real-world examples of DeepFakes (new types of manipulation, different compression schemes, etc.) and challenges related to user experience and explainability of the results.




\vspace{0.3cm}

\noindent\textbf{Acknowledgments:} 
This work has been supported by the WeVerify (contract nr. 825297) and AI4Media (contract nr. 951911) projects, partially funded by the European Commission. 
\normalsize


\bibliographystyle{IEEEtran}  
\bibliography{arxiv_v3}

\end{document}